\documentclass[conference]{IEEEtran}
\IEEEoverridecommandlockouts
\usepackage{amsmath,amsfonts}
\usepackage{algpseudocode}
\usepackage{algorithm}
\usepackage{array}
\usepackage[caption=false,font=normalsize,labelfont=sf,textfont=sf]{subfig}
\usepackage{textcomp}
\usepackage{stfloats}
\usepackage{url}
\usepackage{verbatim}
\usepackage{graphicx}
\usepackage{cite}
\usepackage{rotating}
\usepackage{multirow}
\usepackage{amsthm}
\usepackage{amssymb}
\usepackage{colortbl}
\usepackage{booktabs}
\usepackage[rgb,dvipsnames]{xcolor}
\usepackage{bbding}

\definecolor{aliceblue}{rgb}{0.94, 0.97, 1.0}
\definecolor{lightgray}{rgb}{0.95, 0.95, 0.95}
\def\BibTeX{{\rm B\kern-.05em{\sc i\kern-.025em b}\kern-.08em
    T\kern-.1667em\lower.7ex\hbox{E}\kern-.125emX}}
\begin{document}

\title{UniHPR: Unified Human Pose Representation via Singular Value Contrastive Learning}

\author{Zhongyu Jiang$^1$ \quad
Wenhao Chai$^1$ \quad
Lei Li$^2$ \quad
Zhuoran Zhou$^1$ \quad
Cheng-Yen Yang$^1$ \quad \\
Jenq-Neng Hwang$^1$ \vspace{4pt}\\  $^1$University of Washington, $^2$University of Copenhagen \vspace{4pt} \\
{\tt\small \{zyjiang, wchai, zhouz47, cycyang , hwang\}@uw.edu}
, \tt\small lilei@di.ku.dk \\
}

\maketitle

\begin{abstract} 

In recent years, there has been a growing interest in developing effective alignment pipelines to generate unified representations from different modalities for multi-modal fusion and generation. As an important component of Human-Centric applications, Human Pose representations are critical in many downstream tasks, such as Human Pose Estimation, Action Recognition, Human-Computer Interaction, Object tracking, \textit{etc}. Human Pose representations or embeddings can be extracted from images, 2D keypoints, 3D skeletons, mesh models, and lots of other modalities. Yet, there are limited instances where the correlation among all of those representations has been clearly researched using a contrastive paradigm. In this paper, we propose \textbf{UniHPR}, a unified Human Pose Representation learning pipeline, which aligns Human Pose embeddings from images, 2D and 3D human poses. To align more than two data representations at the same time, we propose a novel singular value-based contrastive learning loss, which better aligns different modalities and further boosts performance. To evaluate the effectiveness of the aligned representation, we choose 2D and 3D Human Pose Estimation (HPE) as our evaluation tasks. In our evaluation, with a simple 3D human pose decoder, UniHPR achieves remarkable performance metrics: MPJPE $49.9$mm on the Human3.6M dataset and PA-MPJPE $51.6$mm on the 3DPW dataset with cross-domain evaluation. Meanwhile, we are able to achieve 2D and 3D pose retrieval with our unified human pose representations in Human3.6M dataset, where the retrieval error is $9.24$mm in MPJPE.


\end{abstract}


\section{Introduction}
\label{sec:intro}

As an important component of human-centric applications, human pose representations (HPRs) are critical in many downstream tasks, such as human pose estimation, action recognition, human-computer interaction, object tracking, etc. Recently, aligning text and human pose sequences (human motion)\cite{tevet2022motionclip, zhang2024motiondiffuse} has been widely discovered. However, there are many more data representations that can be used to denote human poses, including images, 2D keypoints, 3D skeletons, mesh models and etc. From the perspective of representation learning, many previous methods have been dedicated to mapping the representation of human pose sequences into the corresponding text space~\cite{tevet2022motionclip, zhang2024motiondiffuse}. On the other hand, in this paper, we propose \textbf{UniHPR}, a \underline{Uni}fied \underline{H}uman \underline{P}ose \underline{R}epresentation learning framework, which aims to align RGB images, 2D and 3D human poses in the shared feature space. In order to evaluate the quality of the proposed learned representation, we choose human pose estimation (HPE) as our evaluation task. By conducting task-specific fine-tuning, UniHPR can achieve the SOTA performance on both 2D and 3D HPE tasks.

\begin{figure}[t]
    \centering
    \includegraphics[width=\linewidth]{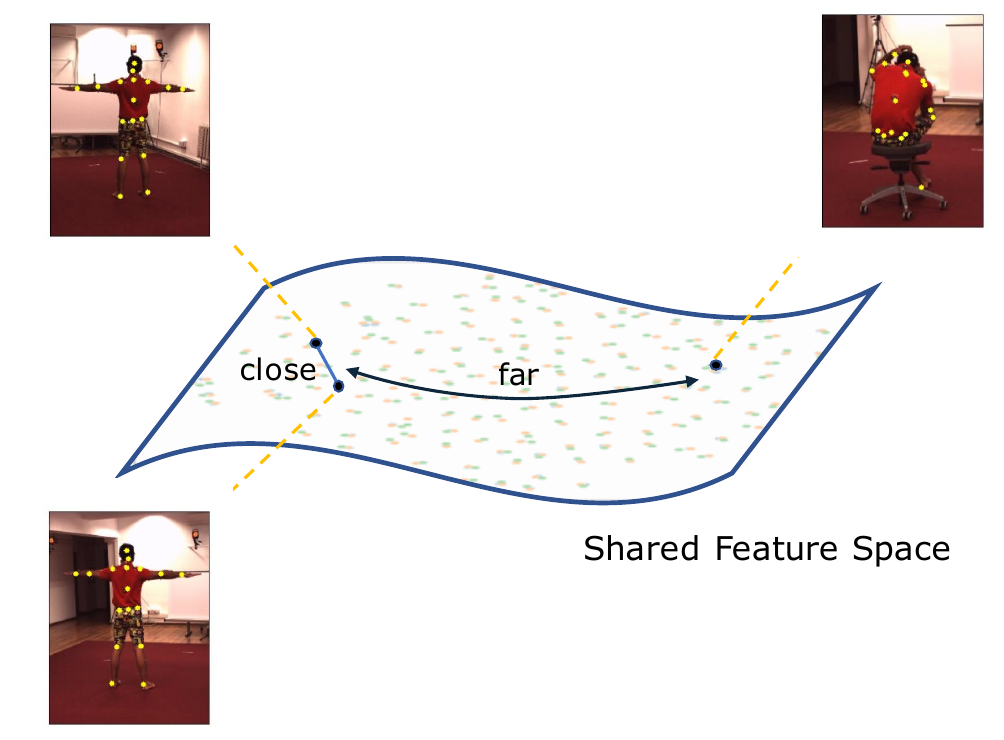}
    \caption{RGB image, 2D and 3D human pose embeddings extracted by corresponding encoders in the shared feature space. After conducting contrastive learning during the pre-training stage, the embeddings extracted from these three different data representations of the same training sample are close to each other and away from other negative samples.}
    \label{fig:intro}
\end{figure}

Learning joint embeddings across more than two data representations (or modalities) is quite challenging. Inspired by Contrastive Language-Image Pre-Training (CLIP)~\cite{radford2021clip}, which proposes to learn aligned visual features with natural language supervisions trained on web-scale image-text paired data. We claim that alignment among RGB image, 2D and 3D human pose representations can also benefit from contrastive learning on large-scale and diverse datasets~(e.g., Human3.6M~\cite{ionescu2013h36m}, MPI-INF-3DHP~\cite{mono-3dhp2017}, etc). 

During the evaluation, UniHPR serves as an encoder with additional downstream task decoders for 2D or 3D HPE. Therefore, the whole pipeline consists of image, 2D and 3D human pose encoders, and 2D and 3D human pose decoders. The embedding features of these three data representations are aligned and shared. To be specific, we first encode the images by HRNet~\cite{wang2020deep}, 2D and 3D human poses by shallow Transformers~\cite{vaswani2017attention} respectively to get the corresponding embeddings, respectively. We then conduct contrastive learning to align the embeddings from these three different data representations of the same training sample in the shared feature space for the unified representation learning. However, aligning embeddings from more than two data representations is challenging, and therefore, we propose a singular value based supervised contrastive learning loss to align three data representations at the same time. After that, during the training stage, we jointly train encoders and decoders with contrastive learning and multi-task learning simultaneously. During inference, since the embeddings are aligned in the same feature space, UniHPR can simultaneously support 2D human pose estimation and 3D human pose estimation, both lifting-based and image-based, in the same pipeline.

Our contributions can be summarised as follows:
\begin{itemize}
    \item We propose the singular value based InfoNCE loss for supervised contrastive learning to effectively align embedding of more than two data representations at the same time.
    \item UniHPR aligns the embedding of Human Pose Representations from three distinctive data representations, i.e., images, 2D and 3D human poses.
    \item With a simple additional diffusion-based decoder, UniHPR achieves SOTA performance on frame-based 3D HPE tasks, e.g., MPJPE $49.9$mm on the Human3.6M dataset with image-3D branch and PA-MPJPE $51.6$mm with 2D-3D branch on the 3DPW dataset for the 3D human pose estimation task.
\end{itemize}

\section{Related Works}
\label{sec:related}

\subsection{Lifting Method for 3D HPE}
 2D-3D lifting~\cite{martinez2017simplebaseline, pavllo2019videopose3d, gong2021poseaug, chai2023global} methods aim to infer 3D human pose under the assistance of the 2D joint detector. Thus, the relations between 2D and 3D human poses have captivated the attention of numerous researchers in computer vision and human motion analysis. Though the internal correspondence is tight, it is rather challenging to align their representations in the embedding space as they contain varying spatial information, and ambiguities in depth may also cause severe one-to-many 2D-3D mappings. 
%
\begin{figure*}[t]
    \centering
    \includegraphics[width=0.9\linewidth]{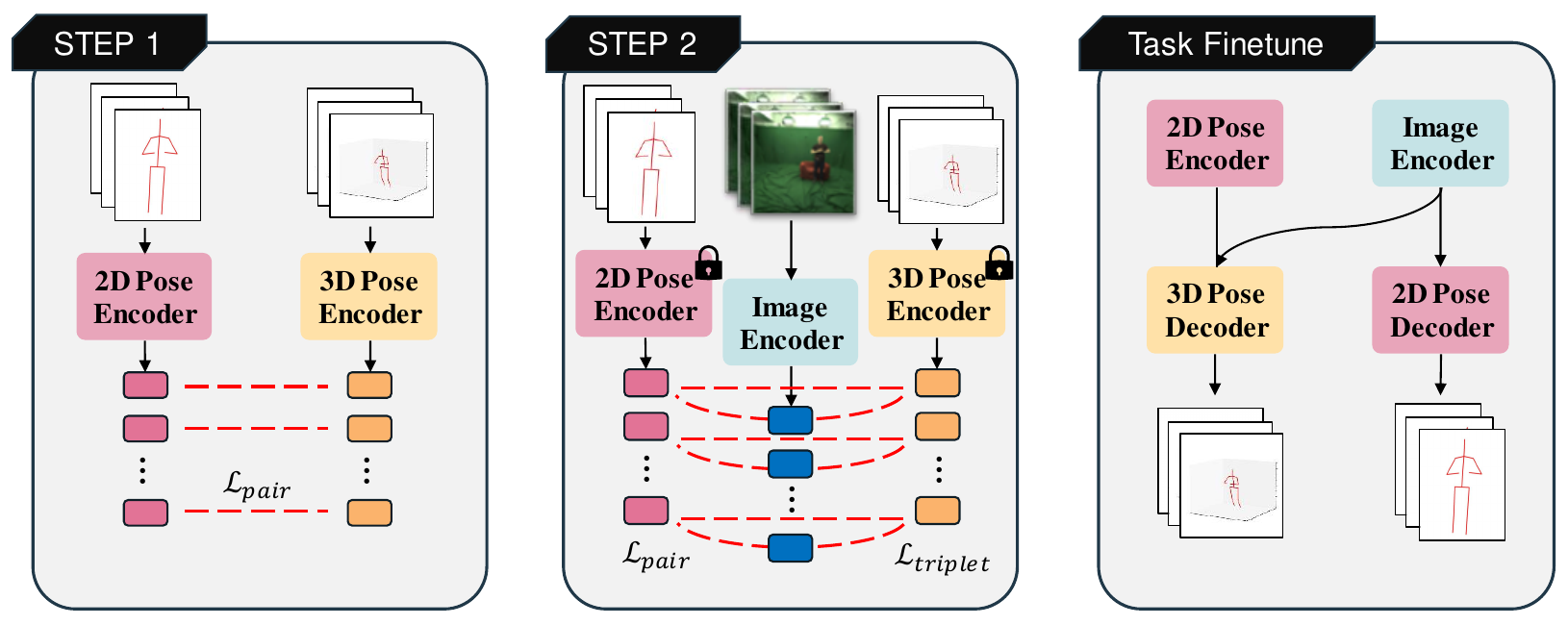}
    \caption{The training scheme of \textbf{UniHPR}. Steps 1 and 2 are representation learning stages, and Task-Specific Finetune is the finetuning stage for any specific task. During Step 1,  the 2D and 3D pose embedding alignment is trained first with $\mathcal{L}_{pair}$, and in Step 2, the image encoder is aligned with frozen 2D and 3D pose encoders via $\mathcal{L}_{pair}$ and $\mathcal{L}_{triplet}$. In the Task-Specific Finetuning stage, encoders and decoders are trained jointly via both contrastive learning, $\mathcal{L}_{pair}$ and $\mathcal{L}_{triplet}$, and task loss.}
    \label{fig:pipeline}
\end{figure*}
\subsection{Image-based Method for 3D HPE}

The other approach for estimating 3D human poses is building an end-to-end network designed to predict the 3D joint coordinates of the poses or SMPL\cite{SMPL:2015} parameters directly from RGB images. Those methods can be categorized into two main classes: heatmap-based~\cite{pavlakos2017coarse, luvizon20182d} and regression-based~\cite{pavlakos2018learning, kolotouros2019spin, kocabas2020vibe, kocabas2021pare, zhang2021pymaf, li2022cliff, li2023hybrik} methods. Following the architecture of 2D human pose estimation, heatmap-based methods generate a 3D likelihood heatmap for each individual joint, and the joint's position is ascertained by identifying the peak within the heatmap. On the other hand, the regression-based methods detect the root location and regress the relative locations of other joints in two branches. In contrast, the SMPL regression methods focus on regressing SMPL parameters from image or video input. Kolotouros et al.~\cite{kolotouros2019spin} propose SPIN, which takes advantage of an optimization-based 3D pose estimation method, i.e., SMPLify~\cite{bogo2016smplify}, to achieve semi-supervised learning on 2D pose only datasets. VIBE~\cite{kocabas2020vibe} utilizes temporal information and a discriminator pretrained on a large 3D pose dataset.


\begin{figure}[t]
    \centering
    \includegraphics[width=\linewidth]{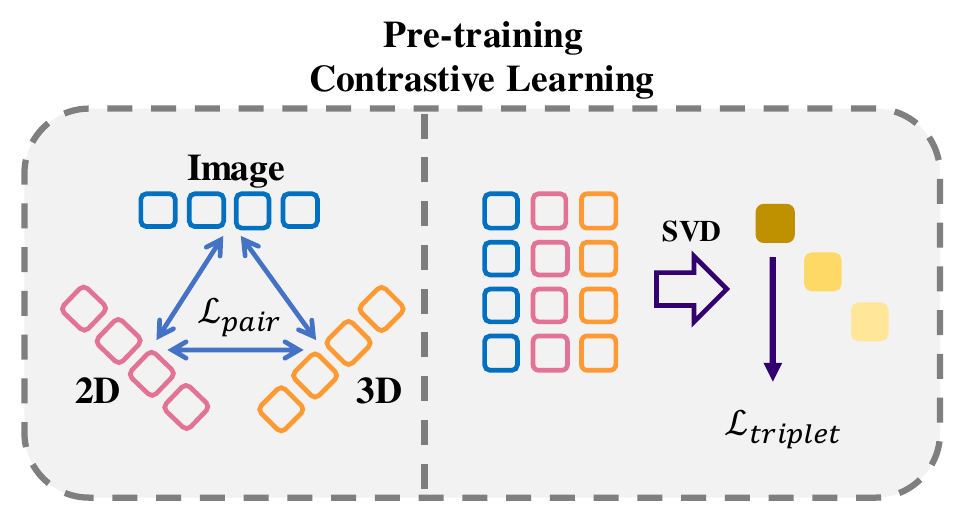}
    \caption{$\mathcal{L}_{pair}$ is applied three times for contrastive learning and the singular value based $\mathcal{L}_{triplet}$ focuses on aligning three representations at the same time.}
    \label{fig:loss}
\end{figure}

\section{Methodology}
\label{sec:method}

We build a unified human pose representation learning pipeline. During training, for any triplet of the cropped human image, $I \in \mathbb{N}^{H \times W \times 3}$, 2D and 3D human poses, $P_{2D/3D} \in \mathbb{R}^{J \times 2/3}$, UniHPR aligns the embeddings from all three representations and utilizes 2D and 3D pose decoders for downstream tasks.

\subsection{Framework Architecture}
\noindent\textbf{Image encoder.}
The extraction of embedding from an RGB image is based on the  HRNet~\cite{wang2020deep}, which is a convolution-based backbone for various visual recognition tasks. We concatenate and flatten the average pooled features from the last stage and pass it through a linear projection layer to obtain a 1-D embedding as our image representation.

\vspace{6pt}
\noindent\textbf{2D/3D pose encoders.}
We adopt two Transformer-based~\cite{vaswani2017attention} encoders to extract the embeddings from 2D and 3D human poses, respectively. We conduct bounding box normalized keypoint-wise patch embedding and retain the spatial information of each keypoint via adding learnable spatial position embedding. Then, the pose tokens prepended with a $\left[CLS\right]$ token and a bounding box token are fed into standard transformer encoder layers, including multi-head self-attention, feed-forward layers, and normalization layers. After that, we use the $\left[CLS\right]$ tokens as 2D and 3D pose embedding, respectively, which effectively aggregates the information of the other tokens and can be regarded as general prior.

\vspace{6pt}
\noindent\textbf{2D and 3D pose decoders.}
We try several different architectures for our task specific decoder, including an MLP, a transformer and a diffusion based model. The diffusion decoder provides the best results in decoding the embedding to generate 2D and 3D human poses. We treat the decoders following the Score Matching paradigm \cite{SMN}. To be specific, the encoded embedding is added with time embedding as well as a data representation token, which indicates the source of the embedding (e.g., from an image, 2D or 3D pose) in the diffusion network as a condition embedding and is used to generate the final 2D and 3D poses. The detailed architectures of all decoders are in the supplemental material.

\subsection{Unified Representation Learning via Contrastive Learning}
During the representation learning stage, we aim to align the embeddings from images, 2D and 3D human poses via the supervised contrastive learning.  Given a batch of data, we have the RGB images, $I \in \mathbb{N}^{B \times H \times W \times 3}$, 2D poses, $P_{2D} \in \mathbb{R}^{B \times J \times 2}$, and 3D poses, $P_{3D} \in \mathbb{R}^{B \times  J \times 3}$, where $B, H, W, J$ are batch size, image height and width, and number of human body keypoints, respectively. The image, 2D, and 3D pose encoders $E_{img}, E_{2D}, E_{3D}$ are trained by maximizing the similarity between image embedding $x_{img} \in \mathbb{R}^{B \times D}$, 2D pose embedding $x_{2D} \in \mathbb{R}^{B \times D}$, and 3D pose embedding $x_{3D} \in \mathbb{R}^{B \times D}$, where $D$ is the dimension of the embedding, which is the same over all three data representations. The most intuitive approach to aligning three embeddings is to apply three pair-wise contrastive losses. 
For embeddings, $x_{\mathcal{S}}, x_{\mathcal{T}}$, from any pair of data representations, the contrastive learning loss is

\begin{equation}
    \mathcal{L}_{pair} = -\log \frac{\exp{(x_{\mathcal{S}} \cdot x_{\mathcal{T}}^{+} / \tau)}}{\sum_{i=1}^{B}\exp(x_{\mathcal{S}} \cdot x_{\mathcal{T},i} / \tau)},
\end{equation}
where $\tau$ is the learnable temperature initialized by $\tau_0$.

However, we found that simply applying three pairwise InfoNCE loss cannot obtain expected embedding similarity across three representations, as shown in the ablation studies in Section~\ref{sec:ablation:loss}. Therefore, we propose a singular value-based InfoNCE loss (Triplet-InfoNCE) to address this issue. 

We stack the embeddings from three representations to build a normalized embedding matrix,  formulated by

\begin{equation}
    \mathcal{M}_x = 
    \begin{bmatrix}
        x_{img} & x_{2D} & x_{3D}
    \end{bmatrix}^T \in \mathbb{R}^{3 \times D}.
\end{equation}

If we apply singular value decomposition (SVD) to this matrix, $M_x = U\Sigma V^*$, the largest singular value, $\sigma_1 = \Sigma_{11}$, is related to the linear correlation of row vectors. Meanwhile, since the embeddings are normalized, the largest singular value should be in $\left[-\sqrt{3}, \sqrt{3}\right]$. Therefore, we can use InfoNCE loss to align any triplet of embeddings by maximizing the $\sigma_1$. However, computing the singular value of a matrix with $3 \times D$, where $3 \ll D$, is time-consuming. Therefore, to accelerate the training procedure, instead of $\sigma_1$, the largest eigenvalue $\lambda_1$ of the matrix $\mathcal{M}_x\mathcal{M}_x^\intercal \in \mathbb{R}^{3 \times 3}$ is the optimization target, since $\lambda_1 = \sigma_1^2$. Therefore, by maximizing the $\lambda_1$ for positive triplets, which contain three embeddings from the same frame, and minimizing the $\lambda_1$ for negative triplets, which contain at least one embedding from a different frame, we are able to align embeddings from three representations jointly.

However, in one minibatch, the number of negative triplets for any positive triplet is $3B^2 - 3B + 1$, and if we use all the negative samples as our denominator in InfoNCE loss, the time consumption is unacceptable. We apply a random sample algorithm to select only $B - 1$ negative triplets for each positive triplet. In this case, the singular value based InfoNCE loss can be formulated as,

\begin{equation}
    \mathcal{L}_{triplet} = -\log \frac{\exp{(\lambda_1^{+} / \tau)}}{\sum_{i=1}^{B}\exp(\lambda_{1i} / \tau)}.
\end{equation}

Overall, our contrastive learning loss is

\begin{equation}
    \mathcal{L}_{cl} = \mathcal{L}_{pair} + \alpha \mathcal{L}_{triplet}.
\end{equation}
where $\alpha $ is the weighted factor.

\begin{figure*}[t]
  \centering
  \subfloat[2D-3D pose]
  {
      \includegraphics[width=0.32\textwidth]{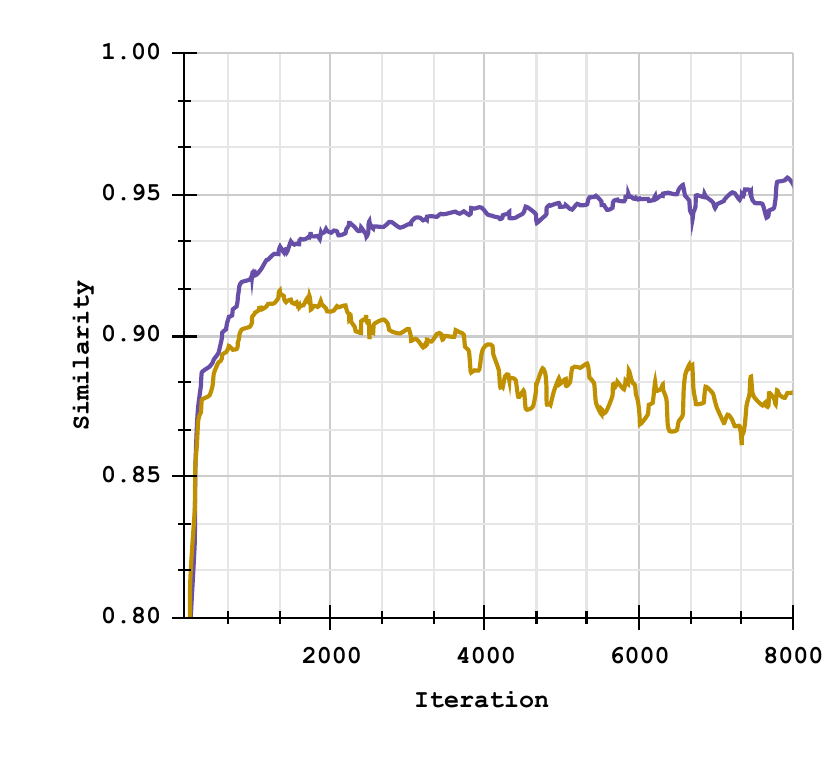}
  }
  \subfloat[image-2D pose]
  {
      \includegraphics[width=0.32\textwidth]{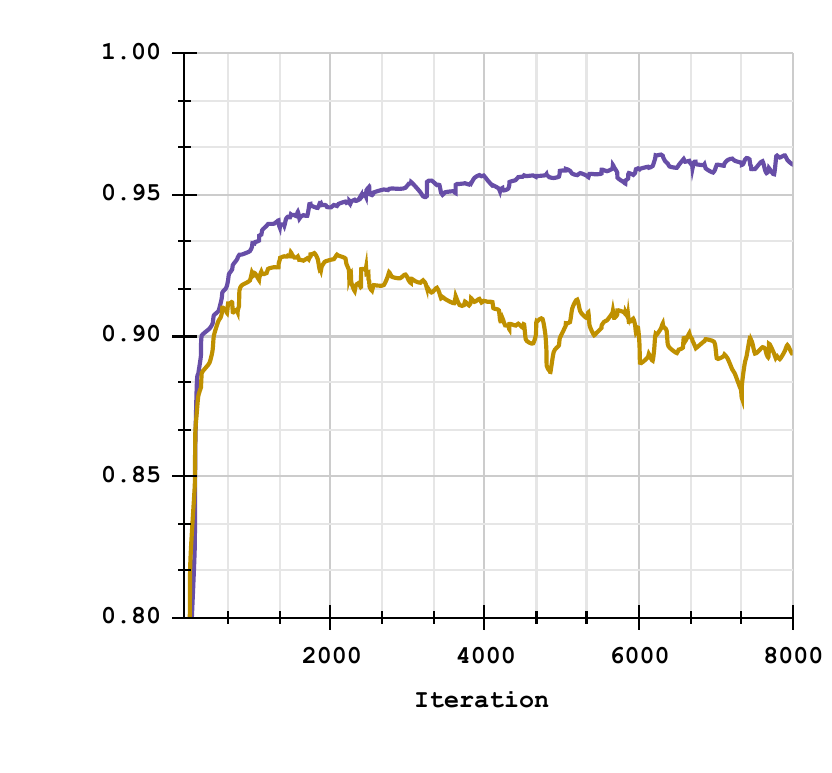}
  }
  \subfloat[image-3D pose]
  {
      \includegraphics[width=0.32\textwidth]{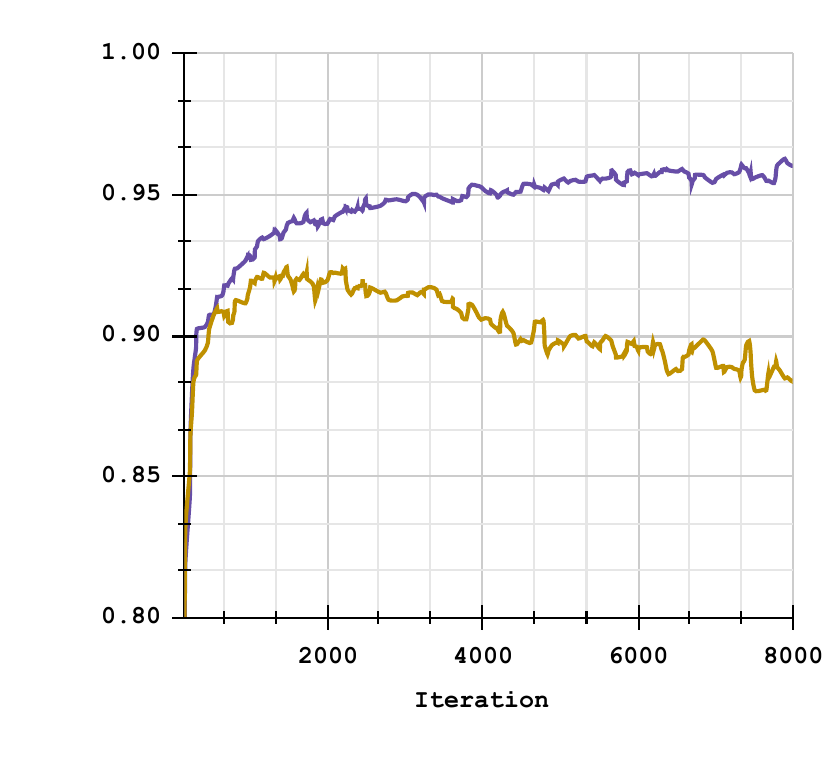}
  }
  \caption{\textbf{Cosine similarities} between different data representations. The yellow line is the one trained only with three pair-wise losses, $\mathcal{L}_{pair}$, and the purple line is the training curve with additional singular value-based InfoNCE loss, $\mathcal{L}_{triplet}$. Our proposed singular value-based InfoNCE loss helps align the feature space.}
  \label{fig:ablation_loss} 
\end{figure*}

\begin{table*}[t]
  \centering
  \caption{\textbf{Image-based 3D HPE} performance on the 3DPW and Human3.6M datasets under MPJPE and PA-MPJPE. \dag~indicates cross-domain evaluation on 3DPW dataset.}
  \resizebox{0.8\linewidth}{!}{
    \begin{tabular}{cl|c|c|cc}
    \toprule
    \multicolumn{2}{c|}{\multirow{2}[4]{*}{Method}} & ~\multirow{2}[4]{*}{Representation}~ & 3DPW & \multicolumn{2}{c}{Human3.6M} \\
\cmidrule{4-6}    \multicolumn{2}{c|}{} &  & ~PA-MPJPE~($\downarrow$)~  & ~MPJPE~($\downarrow$)~  & ~PA-MPJPE~($\downarrow$)~  \\
    \midrule
    \multirow{5}[2]{*}{\begin{sideways}\rotatebox[origin=c]{0}{Temporal}\end{sideways}} & Kanazawa~et al.~\cite{kanazawa2019learning} & SMPL & 72.6 & -     & 56.9 \\
          & Doersch~et al.~\cite{doersch2019sim2real} & SMPL & 74.7  & -     & - \\
          & Arnab~et al.~\cite{arnab2019exploiting} & SMPL & 72.2 & 77.8  & 54.3 \\
          & DSD~\cite{sun2019dsd} & SMPL & 69.5  & 59.1  & 42.4 \\
          & VIBE~\cite{kocabas2020vibe} & SMPL & 56.5 & 65.9  & 41.5 \\
    \midrule
    \multirow{10}[4]{*}{\begin{sideways}\rotatebox[origin=c]{0}{Frame-based}\end{sideways}} & Pavlakos~et al.~\cite{pavlakos2018learning} & SMPL & -     & -     & 75.9 \\
          & HMR~\cite{kanazawa2018hmr} & SMPL & 76.7 & 88.0    & 56.8 \\
          & NBF~\cite{omran2018nbf} & SMPL & -     & -     & 59.9 \\
          & DenseRaC~\cite{xu2019denserac} & SMPL & -     & 76.8  & 48.0 \\
          & SPIN~\cite{kolotouros2019spin} & SMPL & 59.2 & 62.5  & 41.1 \\
          & PARE~\cite{kocabas2021pare} & SMPL & 50.9 & 76.8         & 50.6         \\
          & PyMAF-X~\cite{zhang2023pymafx} & SMPL & \underline{47.1} & 54.2 & 37.2 \\
          & CLIFF~\cite{li2022cliff} & SMPL & \textbf{43.0} & \textbf{47.1} & \textbf{32.7} \\
          \rowcolor{aliceblue}
          & UniHPR-w32\dag~(ours) & Keypoint & 65.8 & 54.5 & 39.5\\
          \rowcolor{aliceblue}
          & UniHPR-w48\dag~(ours) & Keypoint & 64.5 & \underline{49.9} & \underline{35.7} \\
    \bottomrule
    \end{tabular}%
    }
  \label{tab:image-3d}%
\end{table*}%

\begin{table}[t]
  \centering
  \caption{\textbf{Lifting-based 3D HPE} performance on the 3DPW and Human3.6M datasets under MPJPE and PA-MPJPE. The ground truth 2D keypoints are used on 3DPW dataset, while the detected 2D keypoints from CPN are used on Human3.6M dataset. \dag~indicates cross-domain evaluation on 3DPW dataset.
    }
  \resizebox{\linewidth}{!}{
    \begin{tabular}{cl|c|cc}
    \toprule
    \multicolumn{2}{c|}{\multirow{2}[4]{*}{Method}} & {3DPW} & \multicolumn{2}{c}{Human3.6M} \\
\cmidrule{3-5}    \multicolumn{2}{c|}{} & ~PA-MPJPE~($\downarrow$)~  & ~MPJPE~($\downarrow$)~  & ~PA-MPJPE~($\downarrow$)~  \\
    \midrule
    \multirow{4}[4]{*}{\begin{sideways}\rotatebox[origin=c]{0}{Temporal}\end{sideways}} 
    & VideoPose3D~(f=243)~\cite{pavllo2019videopose3d} & 68.0 & 46.8 & 36.5 \\
    & AdaptPose\dag~\cite{gholami2022adaptpose} & 46.5 & - & - \\
    & Li et al.~\cite{li2022exploiting} & - & 43.7 & 35.2 \\
    & MixSTE~\cite{zhang2022mixste} & - & 40.9 & 32.6 \\
    & MPM~\cite{zhang2023mpm} & - & 42.6 & 34.7 \\
    \midrule
    \multirow{4}[4]{*}{\begin{sideways}\rotatebox[origin=c]{0}{Frame-based}\end{sideways}} 
    & SimpleBaseline\dag~\cite{martinez2017simplebaseline} & 89.4 & 62.9 & 47.7 \\
    & SemGCN\dag~\cite{zhao2019semgcn} & 102.0 & 61.2 & 47.7 \\
    & VideoPose3D\dag~(f=1)~\cite{pavllo2019videopose3d} & 94.6 & 55.2 & 42.3 \\
    & PoseAug\dag~\cite{gong2021poseaug} & 58.5 & \underline{52.9} & - \\
    & PoseDA\dag~\cite{chai2023global} & \underline{55.3} & - & - \\
    \rowcolor{aliceblue}
    & UniHPR\dag~(ours) & \textbf{51.6} & \textbf{52.6} & \textbf{39.9}\\
    \bottomrule
    \end{tabular}%
    }
  \label{tab:2d-3d}%
\end{table}%

\begin{table*}[t]
  \centering
  \begin{minipage}[c]{0.48\textwidth}
    \centering
    \caption{\textbf{Quantitative evaluation} of the unified representation. Pose retrieval on Human3.6M test dataset.}
    \resizebox{0.9\linewidth}{!}{
    \begin{tabular}{l|cc}
    \toprule
    Retrieval & ~MPJPE~($\downarrow$)  & PA-MPJPE~($\downarrow$) \\ 
    \midrule
    2D-3D  &   9.2  & 7.1 \\
    Image-3D~ & 10.4 & 7.6 \\
    \bottomrule
    \end{tabular}
    }
  \end{minipage}
  \hspace{0.02\textwidth}
  \begin{minipage}[c]{0.48\textwidth}
    \centering
    \caption{\textbf{Quantitative evaluation} of the unified representation. Image retrieval on Human3.6M test dataset with 1 FPS.}
    \resizebox{0.93\linewidth}{!}{
    \begin{tabular}{l|cc}
    \toprule
    Retrieval & ~Top-1 Acc.~($\uparrow$)  & Top-3 Acc.~($\uparrow$) \\ 
    \midrule
    3D-Image  & 89.2 & 95.6 \\
    2D-Image~ & 95.5 & 97.6 \\
    \bottomrule
    \end{tabular}
    }
    \label{tab:retrieval}
  \end{minipage}
\end{table*}

\begin{table*}[t]
\newcommand{\red}[1]{~\scriptsize{\textcolor{red}{(#1)}}}
\newcommand{\G}[1]{~\scriptsize{\textcolor{OliveGreen}{(#1)}}}
\centering
\caption{\textbf{Ablation study} on UniHPR. Evaluated on Human3.6M dataset. $\mathcal{L}_{pair}$ and $\mathcal{L}_{triplet}$ denotes applying those losses on the pre-training stage. $\mathcal{M}$ token means decoders utilize the representation token. We evaluate the performance with additional data from MPI-INF-3DHP dataset as well.}
\resizebox{0.85\linewidth}{!}{
\begin{tabular}{cccc|cc|cc}
\toprule
  & & & & \multicolumn{2}{c|}{GT 2D} & \multicolumn{2}{c}{Image} \\
 $\mathcal{L}_{pair}$ & $\mathcal{L}_{triplet}$ & $\mathcal{R}$ & w. 3DHP~ & ~MPJPE~($\downarrow$)  & PA-MPJPE~($\downarrow$)~ & ~MPJPE~($\downarrow$)  & PA-MPJPE~($\downarrow$) \\ 
\midrule
\multicolumn{4}{l|}{\textit{baseline}} & 41.3 & 31.6 & 91.8 & 68.7\\
\midrule
\checkmark & & & & 60.0\red{+18.7} & 47.5\red{+15.9} & 65.5\G{-26.3} & 51.8\G{-16.9}\\
\checkmark & \checkmark & & & 40.9\G{-0.4} & 31.7\red{+0.1}  & 58.7\G{-33.1} & 44.4\G{-24.3}\\
\checkmark & \checkmark & \checkmark & & \textbf{39.3\G{-2.0}} & \textbf{29.9\G{-1.7}} & 57.5\G{-34.3} & 42.9\G{-25.8}\\	
\checkmark & \checkmark & \checkmark & \checkmark & 41.7\red{+0.4} & 32.6\red{+1.0} & \textbf{54.5\G{-37.3}} & \textbf{39.5\G{-29.2}} \\
\bottomrule
\end{tabular}
}
\label{tab:ablation}
\end{table*}

\subsection{Task-Specific Finetune}
After the representation learning stage, all encoders and decoders are trained jointly. While encoders are trained with $\mathcal{L}_{cl}$, the task losses, $\mathcal{L}_{2D/3D}$, depend on the architectures of decoders. For the diffusion-based decoder, we adopt the loss from the Score Matching Network~\cite{song2020score}, and for the MLP-based decoder, we utilize $L2$ loss.

Therefore, the overall loss in Task-Specific Finetune is

\begin{equation}
    \mathcal{L} = \mathcal{L}_{cl} + \mathcal{L}_{2D} + \mathcal{L}_{3D}.
\end{equation}

During inference, since the embeddings are well-aligned unified human pose representations in the same feature space, UniHPR can utilize the embedding from any representation and estimate 2D or 3D human poses with shared decoders.

\section{Experiments}

\begin{figure*}[t]
    \centering
    \includegraphics[width=0.95\linewidth]{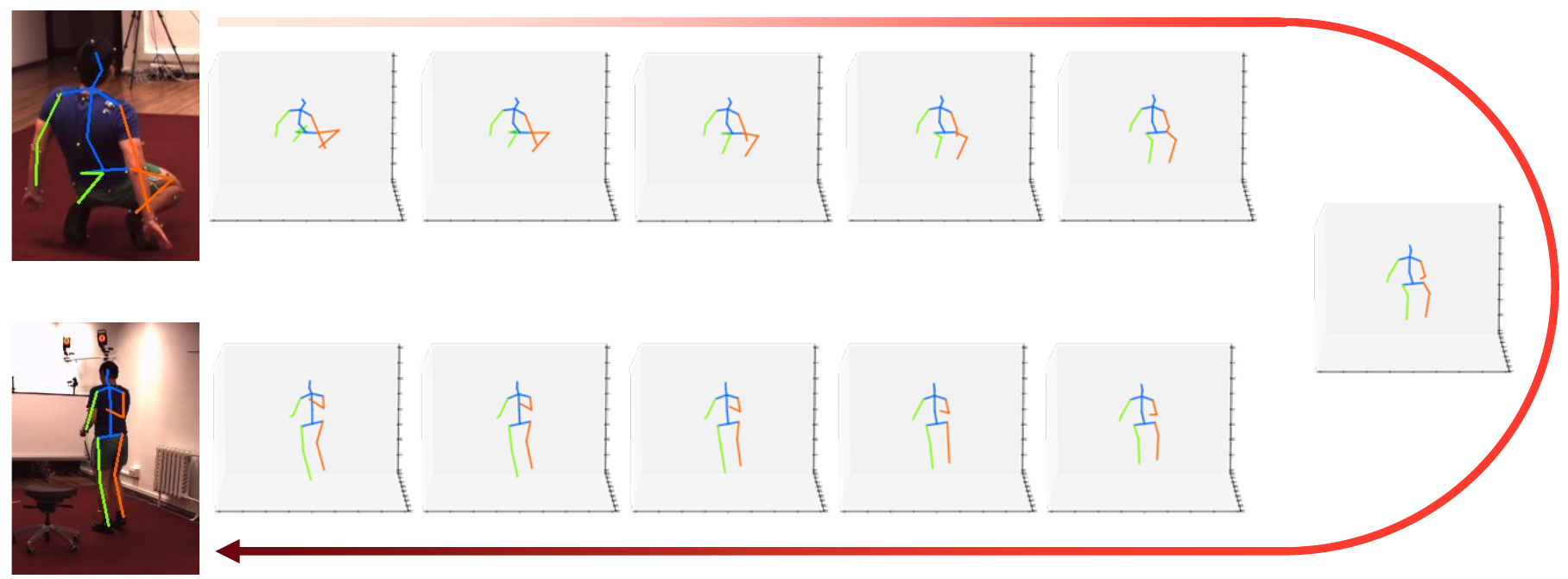}
    \caption{\textbf{Interpolation} of 3D human pose representations in Human3.6M dataset.}
    \label{fig:interpolate}
\end{figure*}

\subsection{Implementation Details}

We implement our proposed framework using PyTorch~\cite{paszke2019pytorch} on a single NVIDIA A100/80G GPU. The representation learning includes two steps: (1) 2D-3D alignment; (2) Image-2D-3D joint alignment (see Fig. 2); followed by a task-specific finetuning stage. In the first step of representation learning, the batch size is $2048$, $\tau_0 = 1/14$, and $\tau \in \left[1/100, 10^4\right]$, while in the second step, the batch size is $180$, $\tau_0 = 1/5$, and $\tau \in \left[1/10, 10^4\right]$. During the multi-task training steps, encoders and decoders are trained together with the batch size being $180$, $\tau_0 = 1/5$, and $\tau \in \left[1/5, 10^4\right]$. For the weight of triplet contrastive loss, $\mathcal{L}_{triplet}$, $\alpha = 1$. The input image size of the image encoder is $192\times256$. During both of the two steps, we adopt Adam optimizer with a learning rate of $1\times 10^{-4}$. We train UniHPR on Human3.6M~\cite{ionescu2013h36m} and MPI-INF-3DHP~\cite{mono-3dhp2017} datasets and apply ablation study about the performance difference on different training datasets.

\subsection{Datasets and Performance Metrics}
To conduct the quantitative performance evaluation of the proposed UniHPR, we use several widely used 3D human pose datasets to train and evaluate our proposed framework, including Human3.6M~\cite{ionescu2013h36m}, MPI-INF-3DHP~\cite{mono-3dhp2017}, and 3DPW~\cite{von20183dpw}. We train UniHPR on Human3.6M and MPI-INI-3DHP and evaluate it on Human3.6M and 3DPW.

\subsection{Evaluation of the Unified Human Pose Representation}

\noindent\textbf{Quantitative Evaluation of Representation Learning.} To better evaluate the quality of learned unified representations, we conduct Pose and Image Retrieval on Human3.6M dataset. The retrieved 3D human pose or image has the most similar 3D pose or image embedding with the image, 2D or 3D pose representation query. For Image Retrieval task, the FPS is set as $1$. In Table~\ref{tab:retrieval}, 2D-3D Pose Retrieval can achieve MPJPE $9.2$mm and the MPJPE of Image-3D Pose Retrieval is $10.4$mm, and 2D-Image Image Retrieval can achieve Top-1 Accuracy $95.5\%$, which illustrate the unified representations are well aligned in images, 2D and 3D human poses. More visualization is included in the supplementary material.


\vspace{3pt}
\noindent\textbf{Interpolation of the Unified Representations.} Furthermore, UniHPR is capable of interpolating 3D human poses by the corresponding 3D pose representations. As shown in Fig~\ref{fig:interpolate}, by interpolating 3D representations from two different 3D poses, UniHPR generates smooth and realistic 3D poses in between.

\subsection{Evaluation of Human Pose Estimation}

\noindent\textbf{Lifting-based 3D Human Pose Estimation}
We evaluate the performance of lifting-based 3D HPE tasks on Human3.6M and 3DPW datasets. As shown in Table~\ref{tab:2d-3d}, UniHPR archives $51.6$ mm in terms of PA-MPJPE on 3DPW dataset and $52.6$ mm in terms of MPJPE on Human3.6M dataset, which is the state-of-the-art performance. Since UniHPR is not trained on 3DPW, it is a fair comparison with those cross-domain evaluation methods.

\vspace{3pt}
\noindent\textbf{Image-based 3D Human Pose Estimation} As for image-based 3D HPE, we also evaluate the performance on Human3.6M and 3DPW datasets. As shown in Table~\ref{tab:image-3d}, UniHPR respectively achieves $49.9$ mm and $35.7$ mm in terms of MPJPE and PA-MPJPE on Human3.6M dataset, as well as $65.7$ mm of PA-MPJPE on 3DPW dataset. Note that we are the only keypoint-based method in Table~\ref{tab:image-3d}, and all the others are SMPL-based. UniHPR achieves comparable performance regarding the number of model parameters and training data with SOTA methods.



\subsection{Ablation Study}
\label{sec:ablation}
In this section, we conduct extensive ablation studies to investigate the importance of each module in the UniHPR, especially how our proposed singular value based loss, $\mathcal{L}_{triple}$, helps the training and improves the performance. 

\vspace{4pt} 

\vspace{4pt} 
\noindent\textbf{End-to-End training without alignment.}
We claim that feature alignment, i.e., pre-training via contrastive learning, among different representations is the key to success. Therefore, we conduct the ablation studies on skipping the alignment training stages. As shown in Table~\ref{tab:ablation}, alignment improves the image-based 3D HPE performance significantly on the Human3.6M dataset. As shown in table~\ref{tab:ablation}, without the 2-step contrastive learning, the performance gap between lifting and image branches shows that the features are not correctly aligned. Furthermore, the combination of $\mathcal{L}_{triplet}$ and $\mathcal{L}_{pair}$ provides the best performance on both lifting and image branches. 

\vspace{4pt} 
\noindent\textbf{Ablation on representation token, $\mathcal{R}$.} In UniHPR, we design a representation token when using the 3D pose decoder to estimate 3D human poses. The representation token indicates which representation the features derived from either (e.g. image or 2D pose). As shown in Table~\ref{tab:ablation},  consistent improvement is observed in using the representation token among lifting-based and image-based 3D HPE tasks on the Human3.6M dataset.

\vspace{4pt} 
\noindent\textbf{Effectiveness of the $\mathcal{L}_{triplet}$.} As shown in Figure~\ref{fig:ablation_loss}, compared to simply applying three pairwise InfoNCE loss, $\mathcal{L}_{pair}$, the proposed singular value-based InfoNCE loss, $\mathcal{L}_{triplet}$, significantly better aligns the features from different representations. With the help of $\mathcal{L}_{triplet}$, the embedding cosine similarity between different representation does not decrease after around $1500$ iterations and keeps increasing to around $0.95$ in $8000$ iterations. For quantitative evaluation, in Table~\ref{tab:ablation}, without the help of $\mathcal{L}_{triplet}$, three $\mathcal{L}_{pair}$ can only achieve MPJPE $65.5$mm and $60.0$mm for image and keypoint branches, which are $6.8$mm and $19.1$mm more than the jointly trained model. \label{sec:ablation:loss}

\vspace{4pt}
\noindent\textbf{Training with additional data.} As shown in Table~\ref{tab:ablation}, it is noted that the distribution of 2D and 3D pose pairs on 3DHP differs from Human3.6M, which increases the robustness of the lifting-based branch but decreases the performance slightly on Human3.6M, since the model trained with both Human3.6M and 3DHP achieves the best performance on 3DPW. Furthermore, training with additional data boosts the image-based branch by improving the diversity of image data.



\begin{figure}[t]
    \centering
    \includegraphics[width=\linewidth]{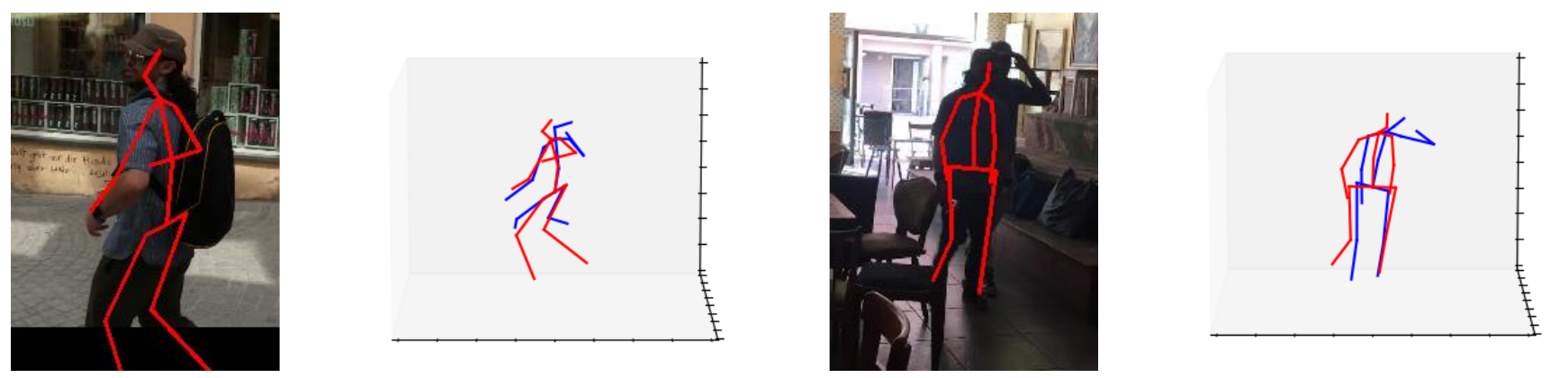}
    \caption{Failure cases of UniHPR. When there is heavy occlusion, our model may estimate the incorrect pose or the pose of a wrong target.}
    \label{fig:failure}
\end{figure}

\vspace{4pt} 
\noindent\textbf{Failure cases.} As shown in Figure~\ref{fig:failure}, the image branch of UniHPR fails in the case of large occlusion or low-quality RGB input scenarios. UniHPR is trained on Human3.6M and MPI-INF-3DHP with only one target per frame and a limited amount of occlusion.

\section{Conclusion}
In conclusion, the UniHPR framework represents a significant step forward in unified human pose representation learning by mitigating the gap between image, 2D and 3D human pose representations. Despite its potential limitations in data and computational requirements, UniHPR sets a promising direction for future research, particularly in improving generalization capabilities and multi-modal representation learning. The framework's achievements on benchmark datasets like Human3.6M and 3DPW justify its potential, paving the way for advancements in applications across multiple domains such as text-to-pose and pose-to-image generation.

\small
\bibliographystyle{IEEEbib}
\bibliography{icme2025references}


\end{document}